\documentclass[11pt,a4paper]{article}
\usepackage[hyperref]{AACL-IJCNLP2020}
\usepackage{times}
\usepackage{latexsym}
\usepackage{dirtytalk}
\usepackage{float}
\usepackage{graphicx}
\usepackage{tabularx}
\usepackage{comment}
\usepackage{adjustbox}
\usepackage{booktabs}
\usepackage{float}
\usepackage{placeins}
\usepackage{stfloats}

\aclfinalcopy 

\title{
An Ensemble Method for Producing Word Representations focusing on the Greek Language}

\author{Michalis Lioudakis \\
  Athens University of \\
  Economics and Business \\ Greece \\
  \texttt{mlioudakis@hotmail.com} \\\And
  Stamatis Outsios \\
  Athens University of \\
  Economics and Business \\ Greece \\
  \texttt{soutsios@aueb.gr} \\\And
  Michalis Vazirgiannis \\
  Athens University of \\
  Economics and Business \\ Greece, \\
  Ecole Polytechnique \\  France \\
  \texttt{mvazirg@aueb.gr} \\}

\date{}

\begin{document}
\maketitle
\begin{abstract}
In this paper we present  a new ensemble  method, Continuous Bag-of-Skip-grams (CBOS), that produces high-quality word representations putting emphasis on the modern Greek language. The CBOS method combines the pioneering approaches for learning word representations: Continuous Bag-of-Words (CBOW) and Continuous Skip-gram. These methods are compared through intrinsic and extrinsic evaluation tasks on three different sources of data: the English Wikipedia corpus, the modern Greek Wikipedia corpus, and the modern Greek Web Content corpus. By comparing these methods across different tasks and datasets, it is evident that the CBOS method achieves state-of-the-art performance.
\end{abstract}

\section{Introduction}

Neural networks have significantly affected  Natural Language Processing (NLP) tasks. One of those tasks is representation learning for words, also known as word embeddings that represent words/tokens in a low dimensional Hilbert space where similarity computations are feasible and enable machine learning algorithms. The main idea behind word embeddings is the distributional hypothesis \cite{harris1954distributional}, which states that the meaning of a word can be captured by the context in which it appears.

Word embeddings are beneficial for most NLP applications increasing the overall  performance and capturing different aspects of similarity among words. Numerous researches have shown these benefits in sequence tagging \cite{ma2016end, lample2016neural} and text classification \cite{kim2014convolutional}. Recently, \citet{qi2018and} have shown that pretrained word embeddings may be a valuable feature in machine translation, particularly in low-resource scenarios.

While living in the NLP era passing from static word representations to dynamic (contextualized) word representations, there are still applications where static word embeddings (word2vec, fastText, GloVe) are used, such as various RNNs/CNNs models. It is also known that in various NLP tasks, using a concatenation of context-aware word embeddings with static word embeddings \cite{peters-etal-2018-deep,akbik2018contextual}  achieves better results.

We propose a new architecture \textbf{Continuous Bag-of-Skip-grams (CBOS)}, aiming to combine the benefits from Skip-gram and CBOW approaches. Our model achieves competitively high accuracy across different tasks compared to the aforementioned models. These results lead to an overall increased performance of the word embeddings. In addition, the CBOS architecture does not increase the computational cost significantly due to its efficient implementation. Thus CBOS can be trained on vast amounts of text corpora within a reasonable time.


The main contributions of our work are: 
\begin{itemize}
  \itemsep0em
    \item Continuous Bag-of-Skip-grams (CBOS), a new ensemble word embeddings method
    \item Two new modern Greek language resources (a dataset for the classification task and a dataset for the NER task)
    \item A comprehensive comparative evaluation of CBOS, CBOW and Skip-gram models trained on three datasets, in two different languages.
\end{itemize}

The rest of the paper is organized as follows: Firstly, section 2 is a brief overview of previous work that has been done on word embeddings and NLP in modern Greek. Section 3 describes the data and tools that were used or produced for the training of our model. In section 4, our proposed CBOS model is explained along with its differences to other popular models. Section 5 presents the evaluation methods used for comparing models in the experimental setup and Section 6 shows the results of the different experiments. Finally, in section 7 we provide conclusions based on the results of the experiments.

\section{Previous Work}

\subsection{Static word embeddings}

Two of the most popular approaches to produce static word vectors are the Skip-gram and the Continuous Bag-of-Words (CBOW) architectures, as implemented in word2vec \cite{Mikolov2013a} and fastText \cite{Bojanowski2017}. The Skip-gram model predicts nearby words given a source word, while the CBOW model, predicts the source word according to its context. The latest version of these models is enriched with subword information in order to overcome some of their shortcomings \cite{Bojanowski2017}.

Even though these two methods produce high-quality word representations, each method achieves the highest accuracy in distinct categories of the word analogy questions. More precisely, the Skip-gram method performs better in semantic categories, while the CBOW method outperforms Skip-gram in syntactic tasks \cite{Mikolov2013a}. Our newly proposed method tries to benefit from both categories in order to increase the overall accuracy.

\subsection{Contextualized word embeddings}

Recent work in the area have shown that contextualized word embeddings outperform traditional word embeddings. This new class of embeddings proposes the production of various representations of each word based on its context, and not a single global representation. Embeddings from Language Models (ELMo) \cite{peters2018deep} is one of these approaches and it is based on the representations obtained by a bidirectional language model. \citet{devlin2018bert} introduced Bidirectional Encoder Representations from Transformers (BERT) which utilizes a deep language model based on a Transformer network.

\subsection{Word embeddings evaluation}

Concerning the comparison of the word representation models, many studies have been focused on word embedding evaluation \cite{ghannay2016word, wang2019evaluating, schnabel2015evaluation}. These studies have examined the intrinsic quality of word embeddings, along with their impact when used as inputs in other NLP application tasks. Thus, we evaluate the word representation methods in two different kind of evaluation tasks: intrinsic and extrinsic evaluation.

\subsection{Resources in modern Greek language}

It is widely known that modern Greek (hereafter simply Greek) resources are limited, especially compared to other rich-resource languages (e.g. English, French, German). Despite the work of \citet{Outsios2018, Outsios2019} that published a large dataset crawled from millions of Greek webpages and an evaluation framework for Greek word embeddings, Greek language continues to be considered as a low-resource language. One of the most recent work in Greek NLP has been published by \citet{koutsikakis2020greek}, where the authors introduced GREEK-BERT, a monolingual BERT-based language model for Greek. Aim of the present work is to enrich the publicly available resources for the Greek language.

\section{Data Sources and Tools}

In this section, we describe the datasets that were used/produced for this research, along with their sources. Furthermore, we present the tools and  libraries that were used for the development of word embeddings models.

\subsection{Wikipedia Corpus}

Wikipedia is the largest, with content in  more than 200 distinct languages, free online encyclopedia. The quality standards followed by authors and the rigorous revisions by editors of the Wikipedia community are ensuring that the articles are of high quality. It has been used in various tasks, among others in information extraction \cite{Wu2010} or word sense disambiguation \cite{Mihalcea2007}.

In this paper, we used the first 10\textsuperscript{9} bytes of the English Wikipedia dump on March 3, 2006 provided by Matt Mahoney\footnotemark. The data is UTF-8 encoded XML consisting primarily of English text. The English Wikipedia corpus contains 243K article titles. The primary preprocessing step was to extract the text content from the XML dumps. For this purpose, the script wikifil.pl was used as published by Matt Mahoney. The final preprocessed file consists from 680MB of text data and 124M words.

\footnotetext{\url{http://mattmahoney.net/dc/textdata.html}}

In addition, the Greek Wikipedia dump from December 2018 was used for training. A few basic preprocessing steps were implemented. These steps included lowercasing of all words and removing punctuation. The finalized text file used for training contains 800MB of text data and 68M words.

\subsection{Greek Web Content Corpus}

Recently, \citet{Outsios2018} have collected and crawled the most extensive Greek corpus available from about 20M URLs with Greek language content. First, the Greek corpus was extracted in Web Archive (WARC) format and then several pre-processing and extraction steps were applied. This process has produced a single uncompressed text which was used by our work. Greek language n-grams were also offered. Some details for the Greek corpus are listed below:

\begin{itemize}
\itemsep0em
  \item Raw crawled text size: 10TB
  \item Text after pre-processing size: 50GB
  \item $|$Tokens$|$: ~3B
  \item $|$Unique sentences$|$: 120M
  \item $|$Unigrams$|$: ~7M
  \item $|$Bigrams$|$: ~90M
  \item $|$Trigrams$|$: ~300M
\end{itemize}

\subsection{New Greek Datasets}
One of our contributions in this work is the production of two new datasets for the text classification and NER tasks for the Greek language. These datasets will be publicly available\footnotemark.
\footnotetext{\url{http://archive.aueb.gr:7000/resources/}}

\subsubsection{Text Classification dataset}\label{clf-dataset}
For the Greek classification task we produced a new dataset from newspaper Makedonia\footnotemark
\footnotetext{\url{http://www.greek-language.gr/greekLang/modern_greek/tools/corpora/makedonia/content.html}}. The full dataset contains 8005 articles from categories like Sports, Reportage, Economy, Politics, International, Television, Arts-Culture, Letters, Opinions etc. For the experiments the top seven categories are selected as a balanced dataset.

\subsubsection{NER dataset}\label{NER-dataset}
For the Greek NER task we produced a new dataset in CoNLL-2003 format, from Spacy's Greek ner.jsonl
\footnotemark.
\footnotetext{\url{https://github.com/eellak/gsoc2018-spacy/blob/dev/spacy/lang/el/training/datasets/annotated_data/ner.jsonl}}

\subsection{FastText Library}

FastText  is an open-source library that allows users to learn text representations and text classifiers. It supports training Continuous Bag-of-Words or Skip-gram models using different loss functions and a variety of tuning parameters. 

Our contribution to fastText Library is the CBOS method that can be used for training. The source code will be made publicly available\footnotemark.
\footnotetext{\url{https://github.com/mikeliou/greek_word_embeddings}}

\section{Proposed Model}

\subsection{Continuous Bag-of-Skip-grams}

The new model, Continuous Bag-of-Skip-grams (CBOS), proposed by this work, is a combination of CBOW and Skip-gram models and was named respectively. The main idea behind CBOS is that, given a word \textit{w} and a context window \textit{c}, the training should capitalize on both training techniques in order to combine their benefits. So we consider two training phases: 

\renewcommand{\theenumi}{\roman{enumi}}
\begin{enumerate}
  \item A phase where \textit{w} is trained by predicting every word in the context window \textit{c} (Skip-gram).
  \item A phase where a bag-of-words is created from all words in the context window \textit{c}, except a randomly selected word \textit{p} which is used for predicting and word \textit{w} which was used for training in the previous phase (CBOW).
\end{enumerate}

Thus, if \textit{D} is the set of correct word-context pairs, the probability functions of the two phases can be defined as follows:

\begin{equation}
    P(D=1|w,c_{1:k})= \prod_{i=1}^{k}{\frac{1}{1+e^{-w \cdot c_{i}}}}
\end{equation}
\begin{equation}
   P(D=1|p,c_{1:k}(p,c_i\neq w))= \\
   \frac{1}{1+e^{-(p \cdot c_{1} + ... + p \cdot c_{k})}} 
\end{equation}

It is essential to  note here, that we selected our proposed CBOS architecture between 6 different implementations, based on accuracy performance in the Greek word analogy task (Section \ref{CBOS-alternatives}).
Furthermore, the CBOS method includes every feature and tuning parameter proposed by \cite{Mikolov2013a,Mikolov2013,Bojanowski2017} as implemented in fastText Library (e.g. subword information, negative sampling).

\begin{figure}[H]
    \begin{center}
        \includegraphics[scale=0.39]{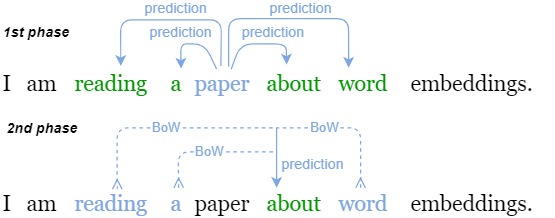} 
        \caption{A visualization of the CBOS model.}
        \label{fig:CBOS}
    \end{center}
\end{figure}

As a working example, consider the sentence \emph{\say{I am reading a paper about word embeddings}} with a window of 2 words before and after the current word. The current word for the first phase of training is \say{paper} and the randomly selected word to predict in the second phase is \say{about}. In the first phase, \say{paper} will make four  predictions, one for each word in the context window (\say{reading}, \say{a},  \say{about}, \say{word}). In the next phase, every word vector, except the one  selected randomly (\say{about}) and  the one used for training in the previous phase (\say{paper}), will be summed in a unique vector and will predict the word \say{about}. This example is illustrated in Figure \ref{fig:CBOS}.

This simple step added to the training of each word seems vital for the improvement of the quality of word embeddings. The additional complexity by this step does not change the complexity class of the algorithm as it appears below (Table \ref{table:tbl2}) in the execution times of the different models. 

As shown, the CBOS model has a few more iterations on the training data than Skip-gram and CBOW models due to the second phase of training. In order to have a fair comparison between the different models in Section \ref{ExperimentalResults}, 
CBOW and Skip-gram models were also trained with double the epoch size (10 instead of 5).

\section{Evaluation}

The quality of our new embedding method was thoroughly evaluated on the basis of restricted lexical semantics tasks, such as scoring word similarity and linear relationships for analogies (intrinsic evaluation). In order to focus on how performance correlates with downstream NLP tasks, two characteristic tasks are selected: a sequence labelling task for word level and a text classification task for sentence level (extrinsic evaluation).

\subsection{Intrinsic Evaluation}

Intrinsic tasks evaluate the quality of word representations generated by an embedding technique. These tasks measure syntactic or semantic relationships between words and, typically, are fast to compute. In this work, three different intrinsic evaluation tasks are selected: word analogy, outlier detection and word similarity.

\subsubsection{Word Analogy}

A common way of evaluating word embeddings is using the vectors produced to predict syntactic and semantic connections like \say{king is to queen as father is to ?}.

\citet{Mikolov2013} were the first to utilize word analogy as a method of creating connections between words, by using the offset of their vectors. The evaluation of word analogy is based upon the observation that simple arithmetic operations in a word vector space can reveal semantic and syntactic relationships between words: given the three words, \emph{a}, \emph{b} and \emph{c}, the task is the identification of the word \emph{d}, so that the relationship \emph{c:d} is the same as the relationship \emph{a:b} \cite{Pereira2016,Turian2010}. The evaluation dataset published by Mikolov and colleagues was used for the evaluation of the English word embeddings in this work.

An evaluation framework for the Greek word embeddings has recently been introduced by \cite{Outsios2019}. This evaluation framework focuses on intrinsic evaluation which evaluates the trained word embeddings using semantic and syntactic analogies and especially in word similarity and word analogy. In this work, for the evaluation of Greek word embeddings, we use the word analogy dataset\footnotemark\footnotetext{\url{http://archive.aueb.gr:8085/files/questions_greek.txt}}.

\subsubsection{Outlier Detection}

Outlier detection task is a relatively new task for evaluating word representations and was proposed by \cite{camacho2016find}. The goal is to distinguish the unrelated word in a group of words. This task evaluates the ability of vector space models to form semantic clusters in order to distinguish the outlier word. Furthermore, \citet{camacho2016find} define Outlier Position Percentage (OPP) which considers the position of the outlier in the group of words ranked by the compactness score.

\subsubsection{Word Similarity}

Word similarity is used for evaluating the distance between word vectors and semantic similarity perceived by humans. The goal of the word similarity task is to evaluate how accurately the human perceived similarity was captured by the word representations. The most common metric used in this evaluation is  cosine similarity.

\subsection{Extrinsic Evaluation}
Extrinsic tasks are used to evaluate the contribution of word representations in the performance of a model in any downstream NLP task. In this work, we chose two different extrinsic evaluation tasks: text classification and named entity recognition.

\subsubsection{Text Classification}
Text classification is one of the most widely used NLP tasks. The goal of this task is to predict the class of the given text. The datasets used for the evaluation of text classification task were: the AG News dataset for the English language\footnotemark and the contributed dataset referenced in Section \ref{clf-dataset} for the Greek language.

\footnotetext{\url{https://www.kaggle.com/amananandrai/ag-news-classification-dataset}}

\subsubsection{Named Entity Recognition}
The named entity recognition (NER) task focuses on locating and classifying information units into pre-defined categories. For example, such categories can be person names, organizations, locations and time expressions.
The datasets used for the evaluation of NER task were: the CoNLL-2003\footnotemark NER dataset for the English language and the contributed dataset referenced in Section \ref{NER-dataset} for the Greek language.

\footnotetext{\url{https://www.kaggle.com/alaakhaled/conll003-englishversion}}

\section{Experimental Results}
\label{ExperimentalResults}

\subsection{Alternative CBOS implementations}\label{CBOS-alternatives}

Before we culminate in the CBOS model proposed earlier, we implemented different versions of CBOS in order to achieve the highest accuracy to the Greek word analogy task. The different versions of CBOS are described below:

\begin{itemize}
  \item \emph{Next-word incremental CBOS}: After the first phase of predictions, the bag-of-words is formed incrementally starting from the first word at the left. After the addition of each word to the bag-of-words, a prediction is made on the next word.
  \item \emph{Central-word incremental CBOS}: The same process as in previous method is followed but, instead of predicting the next word, the prediction is made on the central word of the window.
  \item \emph{Non-random CBOS}: This implementation follows the same steps of CBOS except for the randomly chosen word in the second phase. The chosen word for prediction is the central word.
  \item \emph{Variable context window CBOS}: In the second phase of CBOS, the context window is changed to a random number between 1 and 5. Thus, the bag-of-words could contain different words used for the second phase of training.
  \item \emph{Non-repeated words CBOS}: This method does not add any word to the bag-of-words that is already contained.
\end{itemize}

\begin{table}[!ht]
\begin{center}
\setlength{\tabcolsep}{4pt}
\small
\begin{tabularx}{\columnwidth}{|X|c|c|c|}

      \hline
      \textbf{Model} & \textbf{Semantic} & \textbf{Syntactic} & \textbf{Total} \\
      \hline
      Baseline & \textbf{52.72} & 48.23 & \textbf{50.16} \\
      \hline
      Next-word incremental & 43.35 & \textbf{52.62} & 48.63 \\
      \hline
      Central-word incremental & 9.27 & 36.67 & 24.90 \\
      \hline
      Non-random & 37.96 & 50.60 & 45.17 \\
      \hline
      Variable context window & 48.49 & 47.49 & 47.92 \\
      \hline
      Non-repeated words & 51.12 & 47.65 & 49.14 \\
      \hline

\end{tabularx}
\end{center}
\caption{Accuracy of the different CBOS versions on word analogy task using the Greek Wikipedia Corpus for training.}
\label{table:tbl1}
\end{table}

For the comparison presented in Table \ref{table:tbl1}, the Greek Wikipedia dataset and the default parameters were used for training. For the evaluation, the closest vector is evaluated and the out-of-vocabulary (OOV) words are excluded.

\subsection{Training time}
Before comparing the evaluation scores across the various tasks, in Table \ref{table:tbl2} we present the training time of each model since the computational cost is a critical factor.

\begin{table}[!ht]
\begin{center}
\setlength{\tabcolsep}{4pt}
\small
\newcolumntype{C}{>{\centering\arraybackslash}X}%
\begin{tabularx}{\columnwidth}{|p{2.2cm}|C|C|C|}
      \hline
      \textbf{Model} & \textbf{English Wikipedia} & \textbf{Greek Wikipedia} & \textbf{Greek Web Content}\\
      \hline
      CBOW ep10 & 20m 14.019s & 16m 25.623s & 791m 46.704s \\
      \hline
      CBOW ep5 & \textbf{10m 14.099s} & \textbf{8m 24.453s} & \textbf{399m 7.573s} \\
      \hline
      Skip-gram ep10 & 29m 27.670s & 22m 39.659s & 1395m 20.622s \\
      \hline
      Skip-gram ep5 & 14m 11.977s & 11m 45.747s & 589m 39.222s \\
      \hline
      CBOS ep5 & 21m 10.789s & 16m 55.784s & 810m 43.196s \\
      \hline
\end{tabularx}
 \end{center}
\caption{Training time of the CBOS model and baselines across different datasets.}
\label{table:tbl2}
\end{table}

\subsection{Intrinsic Evaluation}
\subsubsection{Word Analogy}

For the first evaluation, the English Wikipedia dataset was used. The three models were trained using the default parameters provided by the fastText library and were evaluated using the word analogy task for English language \cite{Mikolov2013a}. Only the closest vector (top-1) is considered for a successful prediction. The out-of-vocabulary (OOV) words are excluded. Results are presented in Table \ref{table:tbl3}.

The CBOS model does not achieve the highest accuracy in either the semantic or the syntactic category, but it outperforms the other two models trained with the same epochs in the total score. The CBOW model trained on 10 epochs achieves the best accuracy in the syntactic category. The Skip-gram ep10 model outperforms the other two in the semantic category and the total score but has the worst execution time.

\begin{table*}[!b]
\begin{center}
\setlength{\tabcolsep}{4pt}
\small
\newcolumntype{C}{>{\centering\arraybackslash}X}%
\begin{tabularx}{\linewidth}{|p{2.2cm}|C|C|C|C|C|C|C|C|C|}
      \hline
        & \multicolumn{3}{|c|}{\textbf{English Wikipedia Corpus}} & \multicolumn{3}{|c|}{\textbf{Greek Wikipedia Corpus}} & \multicolumn{3}{|c|}{\textbf{Greek Web Content Corpus}} \\
      \hline
      \textbf{Model} & \textbf{Semantic} & \textbf{Syntactic} & \textbf{Total} & \textbf{Semantic} & \textbf{Syntactic} & \textbf{Total} & \textbf{Semantic} & \textbf{Syntactic} & \textbf{Total} \\
      \hline
      CBOW ep10 
      & 40.21 & \textbf{71.45} & 50.47 
      & 32.71 & 45.16 & 39.81 
      & 21.01 & 55.26 & 43.16
      \\
      \hline
      CBOW ep5 
      & 35.19 & 71.11 & 46.99 
      & 25.14 & 42.93 & 35.29
      & 20.03 & 54.42 & 42.27
      \\
      \hline
      Skip-gram ep10 
      & \textbf{47.26} & 63.80 & \textbf{52.69} 
      & \textbf{58.73} & 41.73 & 49.03
      & \textbf{44.35} & 51.07 & 48.69
      \\
      \hline
      Skip-gram ep5 
      & 43.19 & 61.68 & 49.26 
      & 51.79 & 42.88 & 46.71
      & 41.27 & 52.27 & 48.38
      \\
      \hline
      CBOS ep5 
      & 42.94 & 68.08 & 51.20 
      & 52.72 & \textbf{48.23} & \textbf{50.16} 
      & 41.16 & \textbf{62.39} & \textbf{54.89}
      \\
      \hline
\end{tabularx}
\end{center}
\caption{Accuracy of the CBOS model and baselines on word analogy task using three different datasets for training.}
\label{table:tbl3}
\end{table*}

The next two evaluations used the Greek Wikipedia dataset and the Greek Web Content dataset for the training of the three models. Every model was trained using the default tuning parameters suggested by the FastText framework. The word analogy task for the Greek language \cite{Outsios2019} was used for the evaluation of the closest vector, and the OOV words were not evaluated. The results for the Greek Wikipedia dataset and the Greek Web Content corpus are shown in Table \ref{table:tbl3}.

Concerning the Greek Wikipedia dataset, the CBOS approach achieves the highest accuracy in all categories compared to the models trained on 5 or 10 epochs. The Skip-gram method trained with 10 epochs achieves the highest accuracy in the semantic category, but the CBOS method outperforms the other two methods in the syntactic category and total accuracy.

The results related to the Greek Web Content dataset show that the CBOS method outperforms the other two models in the syntactic category and total accuracy even when they are trained with the double epochs. The Skip-gram ep10 method leads the semantic category.

\subsubsection{Outlier Detection}
We evaluated the models through the outlier detection framework proposed by \cite{camacho2016find}. The results are shown in Table \ref{table:tbl4} below.

\begin{table}[!ht]
\begin{center}
\setlength{\tabcolsep}{4pt}
\small
\newcolumntype{C}{>{\centering\arraybackslash}X}%
\begin{tabularx}{\columnwidth}{|p{2.2cm}|C|C|}
      \hline
      \textbf{Model} & \textbf{OPP Score} & \textbf{Accuracy} \\
      \hline
      CBOW ep10 & \textbf{100.0} & \textbf{100.0} \\
      \hline
      CBOW ep5 & \textbf{100.0} & \textbf{100.0} \\
      \hline
      Skip-gram ep10 & 99.414 & 95.312 \\
      \hline
      Skip-gram ep5 & 99.414 & 95.312 \\
      \hline
      CBOS ep5 & 99.609 & 98.437 \\
      \hline

\end{tabularx}
\end{center}
\caption{Outlier position percentage score and accuracy of the CBOS model and baselines on outlier detection task using the English Wikipedia Corpus for training.}
\label{table:tbl4}
\end{table}

\begin{table*}[!htb]
\setlength{\tabcolsep}{4pt}
\small
\newcolumntype{C}{>{\centering\arraybackslash}X}%
\begin{tabularx}{\linewidth}{|p{2.2cm}|C|C|C|C|C|C|C|C|C|}
      \hline
      \textbf{Model} & \textbf{ESSLI 2c} & \textbf{MEN} & \textbf{WS353} & \textbf{MTurk} & \textbf{Google} & \textbf{MSR} & \textbf{YP} & \textbf{Average} \\
      \hline
      CBOW ep10 & \textbf{0.622} & 0.712 & 0.557 & 0.633 & 0.504 & 0.599 & 0.368 & 0.571 \\
      \hline
      CBOW ep5 & 0.577 & 0.693 & 0.533 & 0.631 & 0.470 & \textbf{0.601} & 0.359 & 0.552 \\
      \hline
      Skip-gram ep10 & 0.6 & 0.737 & \textbf{0.703} & 0.684 & \textbf{0.521} & 0.486 & 0.466 & 0.599 \\
      \hline
      Skip-gram ep5 & 0.577 & 0.733 & 0.691 & 0.684 & 0.457 & 0.461 & 0.436 & 0.577 \\
      \hline
      CBOS ep5 & \textbf{0.622} & \textbf{0.743} & 0.669 & \textbf{0.701} & 0.512 & 0.565 & \textbf{0.475} & \textbf{0.612} \\
      \hline
\end{tabularx}
\caption{Accuracy of the CBOS model and baselines on various word similarity benchmarks using the English Wikipedia Corpus for training.}
\label{table:tbl5}
\end{table*}

The CBOW models reach a perfect score in both metrics regardless of epochs used to be trained, while our proposed CBOS model reaches almost a perfect score as well.

\subsubsection{Word Similarity}

Table \ref{table:tbl5} shows the results of the models evaluated in a range of popular benchmarks used for word similarity. The framework used for this evaluation was developed by \cite{jastrzebski2017evaluate}.

The proposed CBOS model outperforms the other two models in most benchmarks, as well as in average score. The Skip-gram model trained with double epochs achieves the highest score in two datasets (WS353, Google).

\subsection{Extrinsic Evaluation}

\subsubsection{Text Classification}

We divided each dataset in three parts: training, validation and test set. Every experiment was repeated 10 times and the result used is the mean value of Precision, Recall and F1 metrics in the test set. The algorithm used for the text classification task was Linear SVC with its default parameters. 
\begin{table*}[!ht]
\begin{center}
\setlength{\tabcolsep}{4pt}
\small
\newcolumntype{C}{>{\centering\arraybackslash}X}%
\begin{tabularx}{\linewidth}{|p{2.2cm}|C|C|C|C|C|C|C|C|C|}
      \hline
        & \multicolumn{3}{|c|}{\textbf{English Wikipedia Corpus}} & \multicolumn{3}{|c|}{\textbf{Greek Wikipedia Corpus}} & \multicolumn{3}{|c|}{\textbf{Greek Web Content Corpus}} \\
      \hline
      \textbf{Model} 
      & \textbf{Precision} & \textbf{Recall} & \textbf{F1 score} 
      & \textbf{Precision} & \textbf{Recall} & \textbf{F1 score} 
      & \textbf{Precision} & \textbf{Recall} & \textbf{F1 score} \\
      \hline
      CBOW ep10 
      & 69.48 & 69.5 & 69.47
      & 81.91 & 81.52 & 81.64
      & 81.64 & 81.49 & 81.52
      \\
      \hline
      CBOW ep5 
      & 69.38 & 69.40 & 69.38
      & 80.47 & 80.1 & 80.21
      & 81.11 & 81.18 & 81.11
      \\
      \hline
      Skip-gram ep10 
      & \textbf{70.46} & \textbf{70.49} & \textbf{70.47}
      & 82.03 & 81.72 & 81.78
      & 81.37 & 81.12 & 81.19
      \\
      \hline
      Skip-gram ep5 
      & 70.44 & 70.46 & 70.44
      & 81.6 & 81.42 & 81.44
      & 81.82 & 81.45 & 81.58
      \\
      \hline
      CBOS ep5 
      & 70.31 & 70.32 & 70.3
      & \textbf{82.46} & \textbf{82.10} & \textbf{82.18}
      & \textbf{83.37} & \textbf{83.07} & \textbf{83.17}
      \\
      \hline
\end{tabularx}
\end{center}
\caption{Precision, Recall and F1 score of the CBOS model and baselines on Text Classification using Linear SVC algorithm and three different datasets for training.}
\label{table:tbl6}
\end{table*}

The results in Table \ref{table:tbl6} show that the Skip-gram model achieves a slightly higher performance in the evaluation process for the English language.

Concerning the evaluation procedure for the Greek language, the CBOS model outperforms the other two models regardless of training epochs. In particular, the highest difference is achieved when trained with Greek Web Content corpus where our proposed model leads all metrics by 1.5 \% .

\subsubsection{Named Entity Recognition}

For this task we used a bi-LSTM + CRF + chars embeddings model.
We split the datasets in train, validation and test sets, use early-stopping, run every experiment 10 times with random seed and as result use mean value of F1 metric.

\begin{table*}[!ht]
\begin{center}
\setlength{\tabcolsep}{4pt}
\small
\newcolumntype{C}{>{\centering\arraybackslash}X}%
\begin{tabularx}{\linewidth}{|p{2.2cm}|C|C|C|C|C|C|C|C|C|}
      \hline
        & \multicolumn{2}{|c|}{\textbf{English Wikipedia Corpus}} & \multicolumn{2}{|c|}{\textbf{Greek Wikipedia Corpus}} & \multicolumn{2}{|c|}{\textbf{Greek Web Content Corpus}} \\
      \hline
      \textbf{Model} 
      & \textbf{F1-val} & \textbf{F1-test} 
      & \textbf{F1-val} & \textbf{F1-test} 
      & \textbf{F1-val} & \textbf{F1-test}  \\
      \hline
      CBOW ep10 
      & 92.939 & 88.057
      & 71.897 & \textbf{72.335}
      & 75.554 & 76.219
      \\
      \hline
      CBOW ep5 
      & 92.631 & 87.628
      & 70.868 & 70.803
      & \textbf{75.792} & 76.777
      \\
      \hline
      Skip-gram ep10 
      & 92.414 & 87.647
      & 71.897 & \textbf{72.335}
      & 72.407 & 71.949
      \\
      \hline
      Skip-gram ep5 
      & 92.382 & 87.393
      & 72.007 & 72.059
      & 73.054 & 72.104
      \\
      \hline
      CBOS ep5 
      & \textbf{93.055} & \textbf{88.977}
      & \textbf{72.220} & 71.686
      & 74.943 & \textbf{77.145}
      \\
      \hline
\end{tabularx}
\end{center}
\caption{F1 score in validation and test set of the CBOS model and baselines on Named Entity Recognition using three different datasets for training.}
\label{table:tbl7}
\end{table*}

The results in Table \ref{table:tbl7} show that the CBOS model outperforms CBOW and Skip-gram models that have been trained with the same or double number of epochs on the English Wikipedia corpus.

Trained on the Greek Wikipedia Corpus, the CBOS model is between CBOW and Skip-gram models that have been trained with the same number of epochs. Concerning the Greek Web Content Corpus, the CBOS model slightly outperforms CBOW and greatly outperforms the Skip-gram model.

A useful observation is that using F1-val score, where early stopping is used, CBOS is best overall in both English and Greek Wikipedia datasets.

\section{Conclusions}

This paper aimed at producing  high-quality word embeddings mainly for the Greek language, devising a new embedding method, the Continuous Bag-of-Skip-grams (CBOS). CBOS combines the benefits of the CBOW and Skip-gram approaches introduced in \cite{Mikolov2013}. Because of its neat implementation, CBOS  does not increase the computational cost of the training phase.

We presented in Section \ref{ExperimentalResults} that there are particular tasks where CBOW outperforms Skip-gram and vice versa. Since there is no one method that outperforms all others in all tasks, we strongly recommend the use of CBOS. It is evident that CBOS achieves a high performance in every task and the highest performance in most cases regardless of the training epochs.

Additionally, we believe that the CBOS method achieves higher performance in Greek language than English due to their different linguistics aspects \cite{lascaratou1998basic, greenberg1963universals, tzartzanos1963}. For instance, Greek language tends to have longer and more complex sentences. Furthermore, the word order is one of the most important differences of the two languages. The English language is more structured, while the Greek language is more fluid. In conclusion, the CBOS method tends to learn more accurate word representations when the language is more complex and less structured.

The future work of this research could include a more extensive research on CBOS alternatives, which will be based on other criteria than the Greek word analogy task. For instance, the impact of the new embeddings used in an extrinsic task could be considered for comparison. This fine-tuning on the production of word embeddings could lead to significant improvements in extrinsic evaluation tasks.

\section*{Acknowledgments}
We would like to thank Prof. Ion Androutsopoulos for the useful discussions and suggestions.

\bibliography{anthology,aacl-ijcnlp2020}
\bibliographystyle{acl_natbib}

\end{document}